\documentclass[]{spie}  %>>> use for US letter paper
\usepackage{amsmath,amsfonts,amssymb}
\usepackage{graphicx}
\usepackage{multirow}
\usepackage[colorlinks=true, allcolors=blue]{hyperref}
\usepackage[tight,normalsize,sf,SF]{subfigure}

\usepackage[printwatermark]{xwatermark}
\usepackage{xcolor}
\usepackage{graphicx}
\usepackage{lipsum}

\title{Crowdsourcing Lung Nodules Detection and Annotation}

\author[1]{Saeed Boorboor}
\author[1]{Saad Nadeem}
\author[1]{Ji Hwan Park}
\author[2]{Kevin Baker}
\author[1]{Arie Kaufman}
\affil[1]{Department of Computer Science, Stony Brook University, Stony Brook, NY, USA}
\affil[2]{Department of Radiology, Stony Brook Medicine, Stony Brook NY, USA}

\begin{document}
\maketitle

\begin{abstract}
We present crowdsourcing as an additional modality to aid radiologists in the diagnosis of lung cancer from clinical chest computed tomography (CT) scans. More specifically, a complete workflow is introduced which can help maximize the sensitivity of lung nodule detection by utilizing the collective intelligence of the crowd. We combine the concept of overlapping thin-slab maximum intensity projections (TS-MIPs) and cine viewing to render short videos that can be outsourced as an annotation task to the crowd. These videos are generated by linearly interpolating overlapping TS-MIPs of CT slices through the depth of each quadrant of a patient's lung. The resultant videos are outsourced to an online community of non-expert users who, after a brief tutorial, annotate suspected nodules in these video segments. Using our crowdsourcing workflow, we achieved a lung nodule detection sensitivity of over 90\% for 20 patient CT datasets (containing 178 lung nodules with sizes between 1-30mm), and only 47 false positives from a total of 1021 annotations on nodules of all sizes (96\% sensitivity for nodules$>$4mm). These results show that crowdsourcing can be a robust and scalable modality to aid radiologists in screening for lung cancer, directly or in combination with computer-aided detection (CAD) algorithms. For CAD algorithms, the presented workflow can provide highly accurate training data to overcome the high false-positive rate (per scan) problem. We also provide, for the first time, analysis on nodule size and position which can help improve CAD algorithms.
\end{abstract}

\section{Introduction}
Lung cancer is the leading cause of cancer-related deaths in both women and men around the world. Despite its serious outlook, early screening and detection of pulmonary nodules, the precursors of lung cancer, using chest CT scans can dramatically increase the survival rate of lung cancer patients \cite{survivalStats1}. Advancements in CT technology have resulted in higher levels of spatial resolution revealing nodules as small as 1 mm in diameter. However, fundamental to the efficacy of CT screening is the radiologist's ability to detect these nodules within the CT data. While nodules with diameters larger than 10mm can be accurately and rapidly detected, nodules smaller than 10mm in diameter are more difficult and time consuming to detect. These smaller nodules have been found to be critical in early lung cancer detection. According to the National Lung Screening Trial, 35\% of lung cancers patients had nodules that were less than 10mm in size \cite{NLST}, and Wisnivesky et al. similarly showed that smaller the size of the detected nodule, the higher the probability of stage-1 lung cancer~\cite{wisnivesky}.

Nodule detection and chest CT analysis is dependant on the interpreting radiologist and factors such as fatigue, human perception error, image quality and noise, and turn-around time expectations can increase the likelihood of nodule misreads~\cite{girvin}. Moreover, the requirement to search for nodules on every clinical chest CT scan, including those acquired in emergent setting, has resulted in an increase in the volume of scans and thus the number of CT images that require radiologist review. To give a perspective of the complexity involved in interpreting a single chest CT scan, which on average consists of 300 slices with approximately 260,000 pixels/slice: finding a 5mm nodule means detecting a lesion in 5/10,000th of the slice's image area amidst the complex structure and anatomy of the lungs in the background and the nodule's faint attenuation within the depth of the slices. Double reading, computer-aided detection (CAD) and visualization techniques have been proposed to facilitate the screening and expedite the decision-making process for the radiologists. 

Computer-aided detection (CAD) algorithms have been developed and proposed as a primary reader to identify lesion candidates; as a secondary check to identify possible missed lesions, which the radiologist failed to identify, or as a concurrent reader~\cite{bogoni2005computer,mani2004computed}. Specificity is usually sacrificed to increase the CAD sensitivity ~\cite{suzuki2012review}, resulting in an increase of false positives and hence an increase in the interpretation time of the radiologist (in accepting/rejecting the findings). 
A review of CAD false positive results showed that the vast majority were easily dismissed by readers ~\cite{dachman2010effect}. In spite of significant advances in computational algorithms, many problems still require manual intervention, but often the size of the data or the problem is so huge that intervention by a single or a few individuals becomes impractical. Therefore, beyond image-based visualization techniques, CAD and double reading modalities are not extensively used in clinical practice due to limited resources, cost-effectiveness, and lack of general applicability.
%
%Even though CAD systems are useful tools for eventual diagnosis, commercial CAD algorithms have to deal with the unavoidable trade-off between sensitivity and specificity. Therefore, beyond image-based visualization techniques, CAD and double reading modalities are not extensively used in clinical practice due to limited resources, cost-effectiveness, and lack of general applicability.

In this work, we investigate the use of crowdsourcing as an additional modality to aid radiologists in lung nodule detection. Crowdsourcing is a trending modality where cognitive tasks, such as image and video annotation, are outsourced to a pool of untrained individuals from an online community. Recent studies on crowdsourcing has shown promising results, especially in the medical domain, with applications in polyp and false positive detection in virtual colonoscopy \cite{park,summers2012,summers2011}, annotation and reference correspondence generation in endoscopic images \cite{maier}, disease classification \cite{kwitt}, anatomy measurement in CT scans \cite{cheplygina}, etc. To leverage the collective intelligence of the crowd in our context (lung nodule detection and annotation), we couple the concept of maximum intensity projection (MIP) and cine viewing to minimize the nodule miss rate. More specifically, we render videos of overlapping thin-slab MIPs (TS-MIPs) of CT slices through each half of a patient's left and right lungs, and present these to the crowd along with an annotation tool. Our results show that the crowd was able to detect nodules of all sizes with an overall 90\% sensitivity on the Lung Image Database Consortium (LIDC)\cite{lidc} of 20 randomly-selected patient datasets from a set of 1000, with a total of 178 nodules manually annotated by 5 expert radiologists. Moreover, the use of minimal and automated preprocessing steps in our framework allows for rapid scaling of our crowdsourcing application to hundreds of thousands of lung datasets without any overhead.

%Lung cancer is the leading cause of cancer-related deaths in both women and men around the world. Despite its serious outlook, early screening and detection of pulmonary nodules, the precursors of lung cancer, using chest CT scans can dramatically increase the survival rate of lung cancer patients. Nodule detection and chest CT analysis is dependant on the interpreting radiologist and factors such as fatigue, human perception error, image quality and noise, and turn-around time expectations can increase the likelihood of nodule misreads. Moreover, the requirement to search for nodules on every clinical chest CT scan, including those acquired in emergent settings, has resulted in an increase in the volume of scans and thus the number of CT images that require radiologist review. Double reading, computer-aided detection (CAD) and visualization techniques have been proposed to facilitate the screening and to expedite the decision-making process for the radiologists. Even though CAD systems are useful tools for eventual diagnosis, current CAD algorithms have high false positive rates per scan making them less useful for lung nodule detection. Therefore, beyond image-based visualization techniques, CAD and double reading modalities are not extensively used in clinical practice due to limited resources, cost-effectiveness, and lack of general applicability.

\begin{figure}[t!]
  \includegraphics[width = \textwidth]{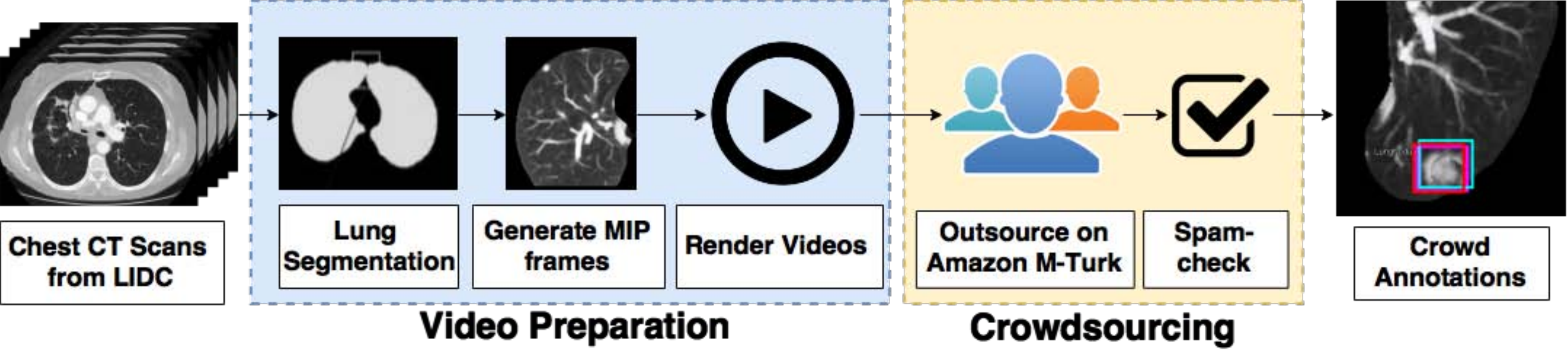}
  \vspace{-5mm}
  \caption{ A pipeline of our proposed experimental study: patient chest CT scans are acquired from the LIDC; overlapping TS-MIP videos of segmented lungs are rendered and outsourced to the crowd for lung detection and annotation}
  %\vspace{2mm}
\end{figure}

%\vspace{-2mm}
\section{METHOD}
\textbf{Dataset.} The crowdsourcing data consists of 20 anonymized chest CT patient scans, from the publicly-available LIDC database, containing 178 nodules covering the range of diameters recommended by the Flieshner Society guidelines~\cite{guidelines}, as shown in Table~\ref{tab:overall-stats}. For establishing ground truth, these scans include a metatdata of the nodules' position in each scan, manually annotated by 5 expert radiologists \cite{lidc-nodule_list}. The nodule distribution in the 20 patient datasets has a mean of 6 nodules per patient and a standard deviation of 3.98 nodules per patient. The distribution of 178 nodules with regards to size was as follows: 91 nodules were 4mm or less, 30 were $>$4mm and $\leq$6mm, 18 were $>$6mm and $\leq$8mm, 15 were $>$8mm and $\leq$10mm, and 24 were larger than 10mm in diameter. From the 20 datasets, 18 were randomly selected and the remaining 2 were extreme patient cases - one with 38 nodules and another with just 1 nodule. The rationale behind selecting these extreme cases was two-fold: (1) to study the sensitivity of the crowd in tasks where they can potentially come across a scan with extraordinary number of nodules in close proximity to each other, and (2) to study the number of false positives in the case where there was only one nodule in the scan.

\begin{figure}[ht!]
\centering
  \includegraphics[width =0.8\textwidth]{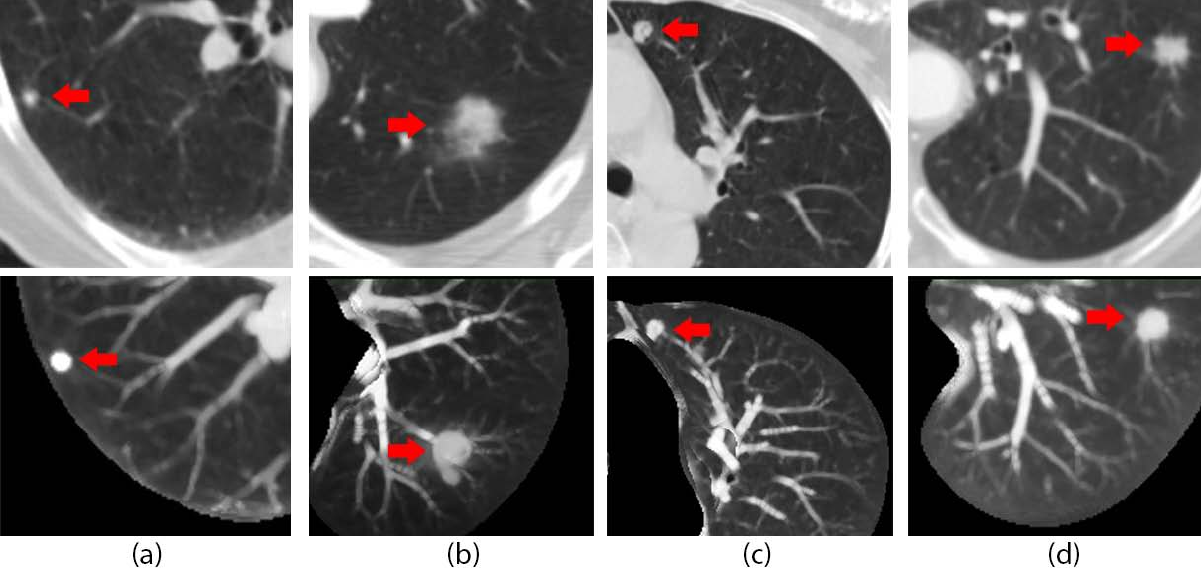}
  \caption{A comparison between a standard CT slice view (top), a TS-MIP of a lung's quadrant as used in this paper (bottom) for: (a) Pleural located nodule; (b) Vessel-attached nodule; (c) Hilar based nodule and (d) Central nodule.}
  \label{ct+mip}
  %\vspace{-3mm}
\end{figure}

%\hspace{-5mm}
\subsubsection{MIP Video Preparation}
Despite the availability of high-resolution CT technology, nodules are still over-looked/misread because of their size, faint attenuation, location within the complex bronchi structures, and proximity to the vessels or abnormal regions such as regions of inflammation. The way in which CT images are reconstructed, visualized and interpreted impacts a radiologist's performance, and thus the goal of reducing perception errors have stimulated researchers to introduce visualization techniques such as MIP, volume rendering, or cine viewing for aiding in the
medical diagnosis.

The goal of this work is to outsource the task of lung nodule detection and annotation to a group of non-expert users. To achieve this goal, we first render a visualization of the CT data that is more conducive to these users and can retain their attention throughout the given task. Prior studies~\cite{rubin,gruden,kawel} have shown an
improvement in lung nodule detection by less-experienced readers when using sliding TS-MIPs in contrast to simple cine viewing of raw CT axial images. Additionally, based on feedback from our in-house user study with 20 non-expert subjects (17 males and 3 females, ages 28$\pm$4), on the effective frame size for cine viewing, we learnt that, screening for nodules throughout the entirety of both lungs simultaneously was a frustrating task, prone to distraction by the anatomy outside the lung parenchyma. With input from expert radiologists, we were able to overcome these challenges by segmenting the lungs, dividing them into four quadrants, and then presenting individual quadrant videos to the crowd for lung nodule detection and annotation.

For preparing these videos, we first automatically segment the lung region
from the CT images using the three steps proposed by Hu et al.\cite{hu}: lung extraction, separation, and smoothing of the lung boundaries. Following the separation step in Hu et al.~\cite{hu}, we divide the lungs into four quadrants and generate running TS-MIPs of 5 consecutive CT slices for each quadrant (Figure 2). These MIP
frames are then linearly interpolated to render a video sequence of 3 frames per second (fps) for each quadrant and the resultant 80 videos (4 quadrant videos per patient dataset) are finally outsourced to the crowd.

% Prior studies have shown an improvement in lung nodule detection by less-experienced readers when using sliding TS-MIPs in contrast to simple cine viewing of raw CT axial images. Additionally, based on feedback from our in-house user study with 20 non-expert subjects (17 males and 3 females, ages 28 $\pm$ 4), on the effective frame size for cine viewing, we learnt that, screening for nodules throughout the entirety of both lungs simultaneously was a frustrating task, prone to distraction by the anatomy outside the lung parenchyma. With input from expert radiologists, we were able to overcome these challenges by (1) automatically segmenting the lung region from the CT scans, (2) dividing the lungs into four quadrants, (3) generating running TS-MIPs of 5 consecutive CT slices for each quadrant (Figure~\ref{ct+mip}), (4) rendering a video sequence of 3 frames per second (fps) for each quadrant, and (5) presenting individual quadrant videos to the crowd for lung nodule detection and annotation.

\subsubsection{Crowd Annotations} 
To reach out to a large and diverse crowd base, we used Amazon Mechanical Turk (MTurk), an internet-based crowdsourcing platform, where cognitive tasks can be distributed to reliable crowd workers at a modest cost. The annotation user interface was integrated into MTurk through an annotation tool called Vatic~\cite{vatic}. For this study, we allotted 10 workers per video on MTurk and awarded \$0.30 to every worker who passed our quality review process, described below.
When a crowd worker voluntarily accepts our advertised task, he/she 
will start with a comprehensive training session, which will describe the
work’s purpose and show how different nodules appear in the chest videos.
%is first introduced to a tutorial about the task. 
The tutorial consists of three parts: (a) short TS-MIP videos of three different nodule types and sizes, (b) a step-by-step instruction set on how to use the interactive annotation tool, and (c) the review
process by which the workers will be paid.

\textbf{The Annotation Tool.} Using the interactive annotation tool, the workers can toggle video controls to watch the short overlapping TS-MIP video segments. While watching the videos, a suspected nodule can be annotated by pausing the video and drawing a bounding box around its region (Figure 3).

\textbf{Quality Review.} To identify spam-workers, who randomly annotate frames for quick monetary reward, we add a small image of a gorilla at a random position in the video; we make sure the gorilla does not overlap with our ground truth lung nodules. Therefore, a minimum requirement for getting paid is the correct annotation of the gorilla in each video. It is important to note that every video has a gorilla, but it may or may not have a nodule; some videos have more than one nodule. In our study, we only consider annotations by workers who have passed this quality review.
%To reach out to a large and diverse crowd base, we used Amazon Mechanical Turk (MTurk), an internet-based crowdsourcing platform, where cognitive tasks can be distributed to reliable crowd workers at a modest cost. The annotation user interface was integrated into MTurk through an annotation tool called Vatic. For this study, we allotted 10 workers per video on MTurk and awarded \$0.30 to every worker who passed our quality review process, described below. When a crowd worker voluntarily accepts our advertised task, he/she is first introduced to a tutorial about the task. The tutorial consists of three parts: (a) short TS-MIP videos of three different nodule types and sizes, (b) a step-by-step instruction set on how to use the interactive annotation tool, and (c) the review process by which the workers will be paid.

%\hspace{-5mm}The Annotation Tool: Using the interactive annotation tool, the workers can toggle video controls to watch the short overlapping TS-MIP video segments. While watching the videos, a suspected nodule can be annotated by pausing the video and drawing a bounding box around its region (Figure~\ref{tool}). We also added a small image of a gorilla in our videos which the workers had to annotate; this helped us distinguish the random clickers from the proper workers; this helped us distinguish random clickers from proper workers.

\begin{figure}
\centering
  \includegraphics[width = 0.8\textwidth]{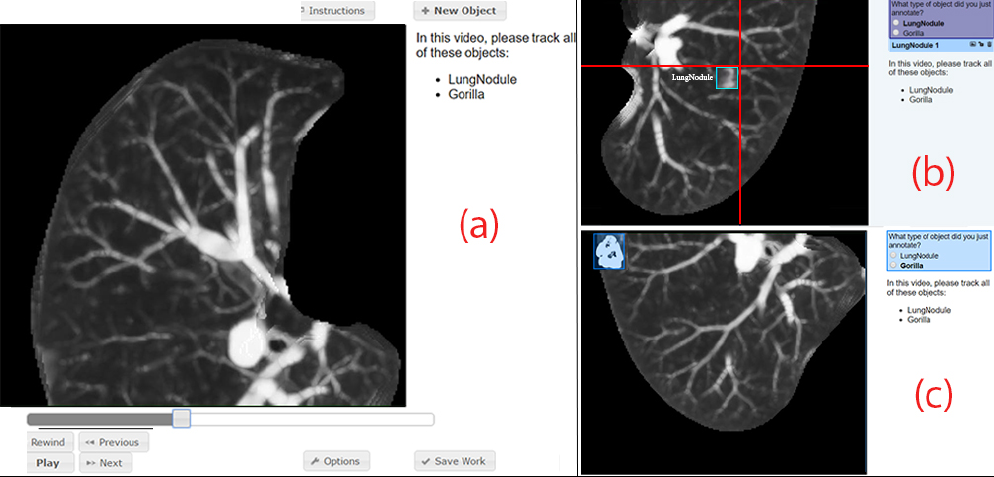}
  \caption{Screen shots of the Vatic annotation tool.(a) The main interface. (b) An example of drawing a bounding box and annotating a nodule. (c) An example of a gorilla that needs to be annotated for quality review.}
  \label{tool}
  %\vspace{2mm}
\end{figure}

%\vspace{-2mm}
\section{Results and Discussions}
For the 80 outsourced videos, we allotted 10 workers per video on MTurk. As
the workers had the liberty to work on more than one video, we collected the
annotations from 143 unique workers. Five crowd attempts were discarded as
spam based on our quality review.
%For the 80 outsourced videos, we allotted 10 workers per video on MTurk. As the workers had the liberty to work on more than one video, we collected the annotations from 143 unique workers. Five crowd attempts were discarded as spam based on our quality review.

The 800 completed tasks contained a total of 1021 annotations. We analyzed the results by comparing the worker annotation with the ground truth annotations. If a worker's bounding box overlapped with an expert radiologist's annotation by more than 60\%, we considered the nodule detected. Out of 178 nodules in the 20 patient CT chest scans, the crowd detected 161 of these nodules, with total of 47 false-positive annotations, resulting in an overall nodule detection sensitivity of 90.4\%. The sensitivities based on the nodule sizes, as shown in Table~\ref{tab:overall-stats}, are as follows: 85.7\% for small-sized nodules ($\leq$ 4mm in diameter), 95.2\% for medium-sized nodules ($>$4mm and $\leq$10mm in diameter), and 95.8\% in case of large-sized nodules ($>$10mm in diameter). Moreover, out of the 10 workers, small-, medium- and large-sized nodules were successfully detected by a mean of 4.2, 8.7, and 9.8 workers respectively, with a median of 4 ($\sigma$ 2.4), 9 ($\sigma$ 1.8), and 10 ($\sigma$ 0.54) respectively. 

\begin{table}[h!]
\centering
\caption{Number of nodules detected by the crowd}
\setlength{\tabcolsep}{7pt}
\label{tab:overall-stats}
\begin{tabular}{|c|c|c|c|}
\hline
\textbf{Nodule Size} & \textbf{No. of Nodules} & \textbf{No. of Nodules} & \textbf{Sensitivity}\\
 (mm in diameter) & \textbf{(Ground Truth)} & \textbf{Detected} & \\
\hline
\hline
\textbf{$\leq$  4}  & 91 & 78 & 85.7\% \\ 
\hline
\textbf{$>$ 4 and $\leq$ 6} & 30 & 28 & 93.3\% \\ 
\hline
\textbf{$>$ 6 and $\leq$ 8} & 18 & 17 & 94.1\% \\ 
\hline
\textbf{$>$ 8 and $\leq$ 10} & 15 & 15 & 100\% \\ 
\hline
\textbf{$>$ 10} & 24 & 23 & 95.8\% \\ 
\hline
\hline
\textbf{Total} & \textbf{178} & \textbf{161} & \textbf{90.4\%}\\
\hline
\end{tabular}
%\vspace{-3mm}
\end{table}

Furthermore, we analyzed the sensitivities based on the nodule location and its attachment within the lung region (Table~\ref{tab:char-stats}). A nodule can be located at the lung periphery or inside the lung region, and it can be attached either to the pleural region (the outer periphery of the lungs), the hilar region (the inner periphery of the lungs that contains the major bronchi, the pulmonary arteries and veins), the vessels, or the central region.

\begin{table}[]
\centering
\caption{Further evaluation of 178 nodules based on location and attachment}
\label{tab:char-stats}
\setlength{\tabcolsep}{3pt}
\begin{tabular}{|c|c|c|c|c|c|c|}
\hline
\textbf{Variable}   &                & \textbf{$\leq$4mm}  & \textbf{4--6mm}         & \textbf{6--8mm}       & \textbf{8--10mm}     & \textbf{$>$10mm} \\
\hline
\hline
\multirow{4}{*}{\textbf{Location}}   & \multirow{2}{*}{\textbf{Peripheral}}     & 42/53         & 14/15       & 8/9       & 8/8         & 13/14     \\  
                            &                                 & (79.3\%)      & (93.3\%)    & (88.9\%)  & (100\%)     & (92.9\%)  \\ \cline{2-7}
                            & \multirow{2}{*}{\textbf{Non-peripheral}} & 36/38         & 14/15       & 9/9       & 7/7         & 10/10    \\
                            &                                 & (94.7\%)      & (93.3\%)    & (100\%)   & (100\%)     & (100\%)  \\
\hline
\hline
\multirow{8}{*}{\textbf{Attachment}} & \multirow{2}{*}{\textbf{Pleural}}        & 27/32         & 12/12       & 5/5       & 6/6         & 8/8    \\
                            &                                 & (84.4\%)      & (100\%)     &(100\%)    &(100\%)      &(100\%) \\ \cline{2-7}
                            & \multirow{2}{*}{\textbf{Vessel}}         & 31/35         & 10/11       & 6/6       & 0/0         & 3/3    \\
                            &                                 &(88.6\%)       &(90.9\%)     &(100\%)    &             &(100\%) \\ \cline{2-7}
                            & \multirow{2}{*}{\textbf{Hilar}}          & 3/7           & 1/2         & 1/2       & 3/3         & 3/4    \\
                            &                                 & (42.9\%)      & (50\%)      & (50\%)    & (100\%)     & (75\%) \\ \cline{2-7}
                            & \multirow{2}{*}{\textbf{Central}}        & 17/17         & 5/5         & 5/5       & 6/6         & 9/9    \\
                            &                                 &  (100\%)      & (100\%)     & (100\%)   &(100\%)      &(100\%) \\
\hline
\end{tabular}
\end{table}

Our analysis for the nodule misreads by the crowd workers (Tables~\ref{tab:overall-stats} and~\ref{tab:char-stats}) shows that apart from a single miss, the crowd managed to detect all the remaining non-peripheral medium- to large-sized nodules and approximately 95\% of the small-sized non-peripheral nodules. However, for the peripheral-located nodules, we observed that for medium- and large-sized nodules, the crowd missed the nodules attached to the hilar region of the lungs. This was due to the similar color intensities of the bronchi, pulmonary arteries and veins that made these nodule seem like part of the structure within the hilar region; a non-expert worker is not trained enough to distinguish between these structures. The same observation holds true for small-sized nodules as only 43\% of the small hilar-attached nodules were detected by the crowd. 

We further noticed that for small-sized nodules, the crowd had a low sensitivity for nodules attached to the vessels or to the pleural region of the lungs. The faint grayish taint of these small-sized nodules, attached to the bright branching vessels, made it difficult for a non-expert user to spot in the MIP videos. Moreover, the artifacts formed at the boundaries of the overlapping MIPs due to the growing and shrinking of the lungs in axial traversal resulted in the occlusion of small nodules attached to the pleural region of the lungs. For our 38-nodule patient dataset, the crowd missed only 2 small-sized nodules and had 5 false positives. For the dataset with only 1 nodule ($>10$mm in diameter), the crowd successfully annotated the single nodule; however, 9 regions were marked as false positives.

\textbf{Limitations and Future Work.} Based on the above analysis, our workflow has two limitations: (1) The MIP artifacts due to the growing and shrinking of the lung axial traversal
occludes the nodules on the periphery and (2) Hilar-attached nodules are poorly
detected. In the future, we will try to overcome these limitations by using volume rendering techniques. Using CT scan data, we can render a three-dimensional (3D) model of the lungs and render high quality MIPs from the 3D model. In addition to artifacts removal, rendering from a 3D volume can improve the MIP-based videos by adding lighting and shading effects to give the vessels and nodules a perception of their shape and structure to the non-expert user. 
For future clinical investigations, we will consider the expertise level of a worker when compiling workers’ answers. For this purpose, we will rank the workers  based on the similarity of
their annotations to other worker annotations. Additionally, the radiologist’s determination will be used to provide a certification level for the workers.

\section{CONCLUSION}
In this work, we presented a crowdsourcing framework for lung nodule detection and annotation to aid radiologists in lung cancer diagnosis. 
We proposed  a protocol to crowdsource lung nodule detection
and leverage collective crowd intelligence by having non-expert workers identify nodule candidates. This protocol
will assist radiologists by substantially reducing reading time and by improving polyp detection sensitivity. These
benefits will allow more people to be screened more effectively, thus reducing the loss of life to lung cancer.
Short videos of lungs were presented to 143 unique untrained workers, who after a brief tutorial, annotated 20 chest CT patient datasets containing 178 pulmonary nodules of varying sizes with a detection sensitivity of over 90\%. 
%Several interesting insights were presented to help build on this work and open up avenues to incorporate crowdsourcing as an additional tool along with CAD, double reading, etc. in the clinical workflow to assist radiologists in the critical task of lung nodule screening. 
We achieved state-of-the-art $>$90\% sensitivity and only 47 false positives from a total of 1021 annotations on nodules of all sizes (96\% sensitivity for nodules$>$4mm); latest 2016 CAD achieved 94\% sensitivity and an average of 8 false positives per scan for nodules$>$4mm. 
%CAD is not used in clinical practice due to the high false-positive rate. The presented workflow can provide highly accurate training data to CAD algorithms to overcome this high false-positve rate problem. We also provided, for the first time, analysis on nodule size and position which can help improve CAD algorithms. 
While this study focuses on evaluating the ability of the crowd, we
anticipate a possible integration with CAD in the future. We foresee that our proposed research on developing and evaluating crowdsourcing for lung nodule annotation will enable faster reading for radiologists, with increased sensitivity mainly due to the crowd acting as a concurrent reader.
Moreover, the use of minimal and automated preprocessing steps in our framework allows for rapid scaling of our crowdsourcing application to hundreds of thousands of lung datasets without any overhead. 

\section{Acknowledgements}
This work has been partially supported by the National Science Foundation grants IIP1069147, CNS1302246, IIS1527200, NRT1633299, and CNS1650499.

% References
% \bibliography{lung} % bibliography data in report.bib
% \bibliographystyle{spiebib} % makes bibtex use spiebib.bst

\end{document}